\documentclass[a4paper,twoside]{article}

\usepackage[utf8]{inputenc}
\usepackage{graphicx}
\usepackage{epsfig}
\usepackage{subfigure}
\usepackage{calc}
\usepackage{amssymb}
\usepackage{amstext}
\usepackage{amsmath}
\usepackage{multicol}
\usepackage{pslatex}
\usepackage{apalike}
\usepackage{fancyhdr}
\usepackage{insticc}
\usepackage[small]{caption}
\usepackage[english]{babel}

\newcommand{\abs}[1]{\ensuremath{\vert #1 \vert}}

\newcommand{\B}[1]{\ensuremath{\mathbf{#1}}}

\def\argmax{\mathop{\rm arg\,max}}

\date{}

\begin{document}
\onecolumn \maketitle \normalsize \vfill

%


\section{\uppercase{Introduction}}
\label{sec:introduction}
\vspace{-0.9cm}

Statistical techniques have been widely used and remarkably successful in automatic speech recognition, machine translation and in natural language processing over the last two decades. This success is due to the fact that this approach is language independent and requires no prior knowledge, only large enough text corpora to estimate probability densities on.
However statistical methods suffer from an intrinsic drawback: they only produce the result which is most likely given training and input data. It is easy to see that this will sometimes not be optimal with regard to human expectations. It is therefore important to be able to automatically evaluate the quality of the result: this can be handled by the different \emph{confidence measures (CMs)} which have been proposed for machine translation.\\

In this paper we introduce new CMs to assess the reliability of translation results. The proposed CMs take advantage of the constituents of a translated sentence: n-grams, word triggers, and also word features.

  \subsection{A Brief Overview of Statistical Machine Translation Principle}
  \label{overviewSMT}
\vspace{-0.2cm}
  In this framework the translation process is essentially the search for the most probable sentence in the target language given a sentence in the source language; let $\B{f}=f_1,..,f_I$ be the source sentence (to be translated) and $\B{\hat e}=e_1,..,e_J$ be the sentence generated by the system (target sentence):

  \begin{equation}
   \B{\hat e} = \argmax_{\B{e}} P(\B{e}|\B{f})
  \end{equation}
  which is equivalent (using the Bayes rule) to:
  \begin{equation}
  \label{smtBaseEq}
   \B{\hat e} = \argmax_{\B{e}} P(\B{e})P(\B{f}|\B{e})
  \end{equation}

  In Equation \ref{smtBaseEq}, $P(\B{e})$ is estimated from a \emph{language model} and is supposed to estimate the correctness of the sentence (``is it a good sentence in the target language ?''), and $P(\B{f}|\B{e})$ is computed from a \emph{translation model} and is supposed to reflect the accuracy of the translation (``does the generated sentence carry exactly the same information than the source sentence ?''). The language model is itself estimated on a large text corpus written in the target language, while the translation model is computed on a bilingual aligned corpus (a text and its translation with line-wise correspondence). The decoder then generates the best hypothesis by making a compromise between these two probabilities.\\
  Of course there are three main drawbacks to this approach: first the search space is so huge that exact computation of the optimum is intractable; second, even if it was, statistical models have inherent limitations which prevent them from being completely sound linguistically; finally, the probability distribution $P$ can only be estimated on finite corpora, and therefore suffers from imprecision and data sparsity. Because of that, any SMT system sometimes produces erroneous translations. It is an important task to detect and possibly correct these mistakes, and this could be handled by confidence measures.

\vspace{-0.4cm}
\section{\uppercase{An Introduction to confidence measures}}
\vspace{-0.2cm}
  \subsection{Motivation and Principle of Confidence Estimation}
\vspace{-0.2cm}
  As said before, SMT systems make mistakes. A word's translation can be wrong, misplaced, or missing. Extra words can be inserted. A whole sentence can be wrong or only parts of it. In order to improve the overall quality of the system, it is important to detect these errors by  assigning a so called confidence measure to each translated word, phrase or sentence. Ideally this measure would be the probability of correctness. An ideal word-level estimator would therefore be the probability that a given word appears at a given position in a sentence; using the notations of Section \ref{overviewSMT} ($e_i$ being the $i$-th word of sentence $\B{e}$), this is expressed by the following formula:
  \begin{equation}
   \textrm{word confidence} = P(correct|i,e_i,\B{f})
  \end{equation}
  and an ideal sentence-level estimator would be:
  \begin{equation}
   \textrm{sentence confidence} = P(correct|\B{e},\B{f})
  \end{equation}

  However these probabilities are difficult to estimate accurately; this is why existing approaches rely on approximating them or on computing scores which are supposed to monotonically depend on them.

  \subsection{State of the Art}
\vspace{-0.2cm}
  Confidence estimation is a common problem in artificial intelligence and information extraction in general \cite{culotta2004cei,gandrabur2006cen}. When it comes to natural language processing, it has been intensively studied for automatic speech recognition \cite{mauclair:t,razik:thesis,guo2004csv}. We find in literature \cite{blatz2003cem,ueffing2004bdr,ueffing2005wlc,uhrik1997cmb,akiba2004umn,duchateau2002csb} different ways of approximating the probability of correctness or of calculating scores which are supposed to reflect this probability.\\

  There are three dominating approaches to estimation of word-level confidence measures for machine translation:

  \begin{itemize}
   \item Estimate words posterior probabilities on the n-best list or word-lattice produced by the decoder (the idea is that correct words will appear frequently).
   \item Use as a confidence estimation the probability that a word in the generated sentence is the translation of a word in the source sentence, by using a translation table.
   \item Transform each word into a vector of numerical features (for example scores coming from different specialised confidence estimators) and train a perceptron to discriminate between ``correct'' and ``incorrect'' classes.
  \end{itemize}

  In \cite{ueffing2004bdr} different original word-level confidence measures are proposed: the word posterior probabilities are estimated from the n-best list, allowing some variation in words positions, and a word's correctness probability is estimated using the translation table generated by an IBM-1 model. Many different confidence measures are investigated in \cite{blatz2003cem}. They are based on source and target language models features, n-best lists, words-lattices, or translation tables. The authors also present efficient ways of classifying words or sentences as ``correct'' or ``incorrect'' by using na\"ive Bayes, single- or multi-layer perceptron.

  \subsection{Our Approach to Confidence Estimation}
\vspace{-0.2cm}
  In the following we will present four different word-level estimators, based on:

  \begin{itemize}
   \item Intra-language mutual information (intra-MI) between words in the generated sentence.
   \item Inter-language mutual information (inter-MI) between source and target words.
   \item An n-gram model of the target language.
   \item A target language model based on linguistic features.
  \end{itemize}

  Mutual Information has been proved suitable for building translation tables \cite{lavecchia:uil} or alignment models \cite{moore}. We use intra-language MI to estimate the relevance of a word in the candidate translation given its context (it is supposed to reflect the lexical consistency). Inter-language MI based confidence estimation gives an indication of the relevance of a translation by checking that each word in the hypothesis can indeed be the translation of a word in the source sentence.
N-gram and linguistic features models estimate the lexical and grammatical correctness of the hypothesis. Finally, because each of these measures targets a specific kind of error, they can be linearly combined in order to obtain a more powerful confidence measure. The weights are optimised on a development corpus.
  Each of these estimators produces a score for every word. This score is then compared to a threshold and the word is labelled as ``correct'' if its score is greater, or ``incorrect'' otherwise. This classification is then compared to a man made reference which gives an estimation of the efficiency of the measures, in terms of error rate, ROC curve and F-measure (see Section \ref{confEval}).

  \subsubsection{Evaluation of the Confidence Measures}
  \label{confEval}
\vspace{-0.25cm}
  As explained before, the CMs are evaluated on a classification task. We manually classified words from 819 sentences generated by MOSES \cite{koehn2007mos} as candidate translations in ``French'' of English sentences extracted from the test corpus of our system (8067 English words, 8816 French words) and ran our classifiers on the same sentences. A word was classified as correct if its score was above a given threshold.
  The results were then compared to the human-made references. We used the following metrics to estimate how well our classifier behaved:

  \paragraph{Classification Error Rate (CER)} is the proportion of errors in classification:
   $$\frac{\textrm{number of incorrectly classified words}}{\textrm{total number of words}}$$
  \paragraph{Correct Acceptance Rate (CAR)} is the proportion of correct words retrieved:
  $$\frac{\textrm{number of correctly accepted words}}{\textrm{total number of correct words}}$$
  \paragraph{Correct Rejection Rate (CRR)} is the proportion of incorrect words labelled as such:
  $$\frac{\textrm{number of correctly rejected words}}{\textrm{total number of incorrect words}}$$
  \paragraph{F-measure} is the harmonic mean of CAR and CRR:
  $$F = \frac{2 \times CAR \times CRR}{CAR + CRR}$$

  These metrics are common in confidence estimation for machine translation \cite{blatz2003cem}. Basically a relaxed classifier has a high CAR (most correct words are labelled as such) and low CRR (many incorrect words are not detected), while a harsh one has a high CRR (an erroneous word is often detected) and a low CAR (many correct words are rejected).\\
  As the acceptance threshold goes from 0 to 1, the classifier becomes harsher: CAR goes from 1 to 0 and CRR from 0 to 1. Therefore we plot CRR against CAR for different thresholds. This tool, very common in information retrieval, is called a \emph{ROC curve}. The ROC curve of a perfect classifier would be the point (1,1) alone, therefore we expect a good classifier to draw a curve as close as possible to the top and right edges of the unit square. This representation is very useful in order to compare the performance of two classifiers: generally speaking, a classifier is better than another if it's ROC curve is always above. In particular, it can be used to quickly visualise the improvement compared to the most naive classifier, which assigns a random score (between 0 and 1) to each word. The ROC curve of such a classifier is the segment going from (0,1) to (1,0), which we plot on our figures. The higher above this line a classifier is, the better. We also plotted on the same diagrams F-measure and CER against CAR.\\



\vspace{-0.4cm}
\section{\uppercase{software and material description}}
\vspace{-0.25cm}
Experiments were run using an English to French phrase-based translation system. We trained a system corresponding to the baseline described in the \emph{ACL workshop on statistical machine translation} \cite{koehn2005epc}. It is based on the state of the art IBM-5 model \cite{brown1994msm} and has been trained on the EUROPARL corpus (proceedings of the European Parliament, \cite{koehn2005epc}) using GIZA++ \cite{och2000gtt} and the SRILM toolkit \cite{stolcke2000ssl}. The decoding process is handled by MOSES.
 The French vocabulary was composed of 63,508 words and the English one of 48,441 words. We summarise in Table \ref{tableCorpSize} the sizes of the different parts of the corpus. This system achieves state of the art performances.

 \begin{table}[ht!]
  \centering
  \scriptsize
  \begin{tabular}{|l|l|l|l|}
  \hline
   & sentences pairs & \multicolumn{2}{c|}{running words} \\
   set & & English & French \\ \hline
  Learning &  465,750 & 9,411,835 & 10,211,388 \\
  Development & 3000 & 75,964  & 82,820 \\
  Test & 1444 & 14,077 & 14,705 \\
  \hline
  \end{tabular}
  \caption{Corpora sizes}
  \label{tableCorpSize}
 \end{table}
 

\vspace{-0.4cm}
\section{\uppercase{Mutual Information based confidence measures}}
\vspace{-0.25cm}
  \subsection{Mutual Information for Language Modelling}
\vspace{-0.25cm}
  In probability theory mutual information measures how mutually dependent are two random variables. It can be used to detect pairs of words which tends to appear together in sentences. Guo proposes in \cite{guo2004csv} a word-level confidence estimation for speech recognition based on mutual information. In this paper we will compute inter-word mutual information following the approach in \cite{lavecchia:uil}, which has been proved suitable for generating translation tables, rather than Guo's.

  \begin{eqnarray}
   \label{imi} MI(x,y) & = & p(x,y) log_2\left(\frac{p(x,y)}{p(x)p(y)}\right) \\
   p(x,y) & = & \frac{N(x,y)}{N} \nonumber \\
   p(x) & = & \frac{N(x)}{N} \nonumber
  \end{eqnarray}
  where $N$ is the total number of sentences, $N(x)$ is the number of sentences in which $x$ appears and $N(x,y)$ is the number of sentences in which $x$ and $y$ co-occur. We smooth the estimated probability distribution, as in Guo's paper, in order to avoid null probabilities:
  \begin{eqnarray}
   N(x,y) & \gets & N(x,y) + C\\
   p(x,y) & \gets & \frac{p(x,y)+\alpha p(x)p(y)}{1+\alpha}
  \end{eqnarray}
  in which $C$ is a non-negative integer and $\alpha$ a non-negative real number. For example, words like \textit{``ask''} and \textit{``question''} have a high mutual information, while words coming from distinct lexical fields (like \textit{``poetry''} and \textit{``economic''}) would have a very low one. Since it is not possible to store a full matrix in memory, only the most dependent word pairs are kept: we obtain a so called \emph{triggers list}.

  \subsection{Confidence Measure based on Intra-Language Mutual Information}
  \label{imiCM}
\vspace{-0.25cm}
  By estimating which target words are likely to appear together in the same sentence, intra-language MI based confidence score is supposed to reflect the lexical consistency of the generated sentence. We computed mutual information between French words from the French part of the bilingual corpus. Table \ref{tableIMI} shows an example of French intra-lingual triggers, sorted by decreasing mutual information.

  \begin{table}[ht!]
  \centering
  \scriptsize
  \begin{tabular}{|lcl|l|}
    \hline
    \textit{first word} & $\rightarrow$ & \textit{triggered word} & mutual information \\
    \hline
    \textit{sécurité} & $\rightarrow$ & \textit{alimentaire} & $ 4.43 \cdotp 10^{-3}$ \\
    \textit{sécurité} & $\rightarrow$ & \textit{étrangère} & $ 4.27 \cdotp 10^{-3} $ \\
    \textit{sécurité} & $\rightarrow$ & \textit{politique} & $ 4.06 \cdotp 10^{-3} $ \\
     & ... &  & \\
    \textit{politique} & $\rightarrow$ & \textit{commune} & $ 1.00 \cdotp 10^{-2} $ \\
    \textit{politique} & $\rightarrow$ & \textit{\'economique} & $ 8.46 \cdotp 10^{-3} $ \\
    \textit{politique} & $\rightarrow$ & \textit{\'etrang\`ere} & $ 7.88 \cdotp 10^{-3} $ \\
    \hline
  \end{tabular}
  \caption{An example of French intra-lingual triggers}
  \label{tableIMI}
  \end{table}

  Let $\mathbf{e} = e_1 .. e_I$ be the target sentence. The score assigned to $e_i$ is the weighted average mutual information between $e_i$ and the words in its context:

  \begin{eqnarray}
  \label{confIMI}
   C(e_i) & = & \frac{\sum_{j=1..I,j\not=i} w(\abs{i-j})MI(e_j,e_i)}{\sum_{j=1..I,j\not=i} w(\abs{i-j})}
  \end{eqnarray}
  where $w()$ is a scaling function lowering the importance of long range dependencies. It can be constant if we do not want to take words' positions into account, exponentially decreasing if we want to give more importance to pairs of words close to each other, or a shifted Heaviside function if we want to allow triggering only within a given range (which we will refer to as \emph{triggering window}).\\
  We also experimented with different kinds of normalisation:
  \begin{itemize}
   \item Beforehand normalisation of the \textbf{triggers list}:
   \begin{equation}\label{normtrigs} MI(x,y) \gets \frac{MI(x,y)}{max_{y'}{MI(x,y')}} \end{equation}
   \item Normalisation with regard to \textbf{norm-}$1$ as in \cite{ueffing2004bdr}:
   \begin{equation}\label{norm1} C(e_i) \gets \frac{C(e_i)}{\sum_{j=1}^I \abs{C(e_j)}} \end{equation}
   \item With regard to \textbf{norm-}$\infty$:
   \begin{equation}\label{norminf}C(e_i) \gets \frac{C(e_i)}{\max_{j=1..I} \abs{C(e_j)}}  \end{equation}
  \end{itemize}

  Tool words like \textit{``the''}, \textit{``of''},...
  tend to have a very high mutual information with all other words thus polluting the trigger list. We therefore ignored them in some of our experiments.\\

\begin{figure}[ht!]
    \centering
    {\includegraphics[width=5.2cm]{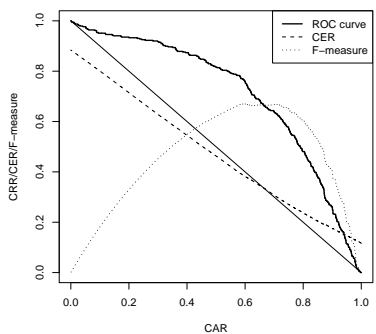}}
    \caption{Intra-MI, no tool words, no normalisation, no weighting nor triggering window.}
    \label{ROCimi1}
    \end{figure}

    Presenting the performances of the confidence measure with all different settings (normalisation, with or without tool words,...) would be tedious. Therefore we only show the settings that yield the best performances. Note that while other settings often yield much worse performance, a few perform almost as well, therefore there are no definite ``optimal settings''. Figure \ref{ROCimi1} shows the ROC curve, CER and F-measure of a classifier based on intra-MI: tool words were ignored, no normalisation was applied, and words positions were not taken into account. Remember that these curves are obtained by assigning a score to each word in the generated sentences, then, for different thresholds between 0 and 1, classifying all these words as correct or incorrect. Each of these thresholds gives a CAR, a CRR and a CER and therefore a point of the curves.

%

    This classifier shows very interesting discriminating power : for a CAR of 50\% the CRR is slightly above 80\% (harsh classifier), and for a CRR of 50\% the CAR is almost 80\% (relaxed classifier). We empirically found that taking word positions into account in intra-MI based confidence measures tends to yield lower performance. We interpret in the following way: intra-language MI reflects lexical consistency of the sentence, but two related words may not be next to each other in the sentence.

  \subsection{Confidence Measure Based on Inter-language Mutual Information}
  \label{cmiCM}
\vspace{-0.25cm}
  The principle of intra-language MI was to detect which words trigger the appearance of an other word in the same sentence. This principle can be extended to pairs of source and target sentences \cite{lavecchia:uil}: let $N_S(x)$ be the number of source sentences in which $x$ appears, $N_T(y)$ the number of target sentences in which $y$ appears, $N(x,y)$ the number of pairs \textit{(source sentence, target sentence)} such that $x$ appears in the source and $y$ in the target, and $N$ the total number of pairs of source and target sentences. Then let us define:
  \begin{eqnarray}
   p_S(x) & = & \frac{N_S(x)}{N} \nonumber \\
   p_T(y) & = & \frac{N_T(y)}{N} \nonumber \\
   p(x,y) & = & \frac{N(x,y)}{N} \nonumber \\
   MI(x,y) & = & p(x,y)log_2\left(\frac{p(x,y)}{p_S(x)p_T(y)}\right)
  \end{eqnarray}
  Guo's smoothing can be applied as in Section \ref{imiCM}.
  One then keeps only the best triggers and obtain a so-called \emph{inter-lingual triggers list}.
  Table \ref{tableCMI} shows an example of such triggers between English and French words, sorted by decreasing mutual information.

  \begin{table}[ht!]
  \begin{center}
  \scriptsize
  \begin{tabular}{|lcl|l|}
    \hline
    \textit{English word} & $\rightarrow$ & \textit{triggered French word} & MI \\
    \hline
    \textit{security} & $\rightarrow$ & \textit{sécurité} & $ 8.03 \cdotp 10^{-2} $\\
    \textit{security} & $\rightarrow$ & \textit{étrangère} & $ 8.55 \cdotp 10^{-3} $ \\
    \textit{security} & $\rightarrow$ & \textit{politique} & $ 6.08 \cdotp 10^{-3} $ \\
      & ... & & \\
    \textit{policy} & $\rightarrow$ & \textit{politique} & $ 2.62 \cdotp 10^{-2} $ \\
    \textit{policy} & $\rightarrow$ & \textit{commune} & $ 3.39 \cdotp 10^{-3} $ \\
    \textit{policy} & $\rightarrow$ & \textit{étrangère}  & $ 2.71 \cdotp 10^{-3} $ \\
    \hline
    \end{tabular}
  \end{center}
  \caption{An Example of Inter-Lingual triggers}
  \label{tableCMI}
  \end{table}

  The confidence measure is then:

  \begin{eqnarray}
  \label{confCMI}
   C(e_i) & = & \frac{\sum_{j=1}^J w(\abs{i-j})MI(f_j,e_i)}{\sum_{j=1}^J w(\abs{i-j})}
  \end{eqnarray}

\begin{figure}[ht!]
     \centering{
     \includegraphics[width=5.2cm]{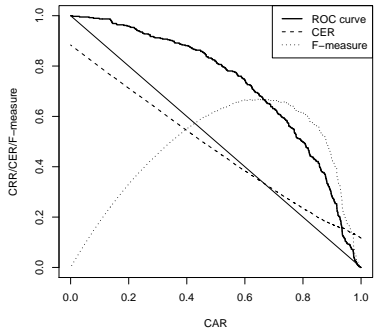}
     }
     \caption{Inter-language MI based CM: tool words excluded, no normalisation, triggering is allowed within a centred window of width 9.}
     \label{ROCcmi2}
     \end{figure}

  We show in Figure \ref{ROCcmi2} the characteristics of such an inter-MI based classifiers. No normalisation whatsoever was applied, and tool words were excluded. This time triggering was allowed within a window of width 9 centred on the word the confidence of which was being evaluated.

    Unlike intra-MI based classifier, we found here that setting a triggering window yields the best performance. This is because inter-language MI indicates which target words are possible translations of a source word. This is much stronger than the lexical relationship indicated by intra-MI; therefore allowing triggering only within a given window or simply giving less weight to ``distant'' words pairs reflects the fact that words in the source sentence and their translations in the target sentence appear more or less in the same order (this is the same as limiting the distorsion, which is the difference between the positions of a word and its translation).

\vspace{-0.4cm}
\section{\uppercase{n-grams based confidence measure}}
\vspace{-0.25cm}
  \subsection{Principle}
\vspace{-0.25cm}
  Remember Equation \ref{smtBaseEq}: the decoder makes a compromise between $P(\B{e})$ (which we will refer to as \emph{language model score}) and $P(\B{f}|\B{e})$ (\emph{translation score}). Because of that, if a candidate $\B{e}$ has a high translation score and a low language model score, it might be accepted as the ``best'' translation. But a low LM score often means an incorrect sentence and therefore a bad translation. This consideration applies on sub-sentence level as well as on sentence level: if the n-gram probability of a word is low, it often means that it is wrong or at least misplaced. Therefore we want to use the language model alone in order to detect incorrect words. We decided to use the word probability derived from an n-gram model as a confidence measure:

  \begin{eqnarray}
    C(e_i) & = & P(e_i|e_{i-1},..,e_{i-n+1})
  \end{eqnarray}

\begin{figure}[ht!]
    \centering
    \includegraphics[width=5.2cm]{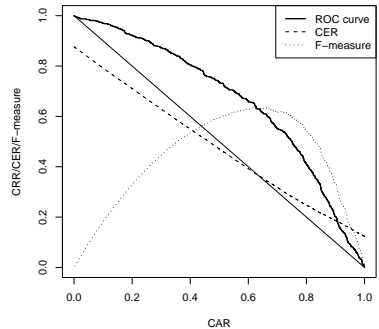}
    \caption{Performance of a 4-grams language model based classifier.}
    \label{ROC4grams}
  \end{figure}

  While intra-language triggers are designed to estimate the lexical consistency of the sentence, this measure is supposed to estimate its well-formedness. Figure \ref{ROC4grams} shows the performances of a 4-grams based classifier. 

  While still showing an interesting discriminating power, it does not perform as well as the best MI-based classifiers: some hypothesis with a low language model score will indeed have already been discarded by the decoder. Also we classify as incorrect only the last word of the n-gram, however a low n-gram score indicates that the sequence (or any word in it) is wrong, rather than only the last word.

\vspace{-0.4cm}
\section{\uppercase{Linguistic features based confidence measure}}
\vspace{-0.25cm}
  \subsection{Principle}
\vspace{-0.25cm}
  Classical language models do not directly take into account tense, gender and number agreement between the different words of the output sentence. We want to specifically target agreement errors: this is why in the following we propose a confidence measure based on linguistic features. For that, we use BDLEX \cite{decalmes:bls} to replace each word by a vector of features \cite{smaili:sfl}. We specifically select three features:

  \begin{itemize}
   \item \textbf{Syntactic class}, for example noun, verb, etc...
   \item \textbf{Tense} of verbs or nothing for other classes.
   \item \textbf{Number and gender} of nouns, adjectives or past participles, \textbf{person} for verbs, nothing otherwise.
  \end{itemize}

\begin{figure}[ht!]
    \centering{
    \includegraphics[width=5.2cm]{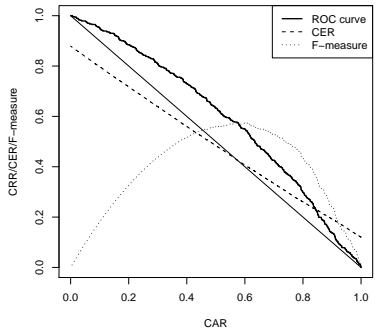}
    }
    \caption{Performance of a classifier based on a 6-grams linguistic features model.}
    \label{ROCfeats6best}
  \end{figure}

  For example the word \textit{\'etait} becomes \textit{V,ii,3S} standing for ``verb, imperfect indicative, 3rd person''. We then train a classical n-gram model on the generated features corpus using the SRILM toolkit and use the n-gram probability as a confidence estimation. In Figure \ref{ROCfeats6best} we print the ROC, CER and F-measures of confidence measures based on a 6-grams linguistic features language model.

    The performances of this CM are rather disappointing, and the CER is particularly terrible. This can probably be at least partially explained by the difficulty of disambiguation (some words belong to different classes, like the French word ``sommes'' which can be a conjugated verb or a plural noun): because we have no information that might allow us to perform a correct choice, it was randomly performed during training of the model, and during sentence evaluation the most likely class (according to the previously trained linguistic feature n-gram model) was chosen. We believe progress could certainly be made by performing smarter disambiguation.

\vspace{-0.4cm}
\section{\uppercase{Fusion of confidence measures}}
\begin{figure}[ht!]
  \centering
  \includegraphics[width=5.2cm]{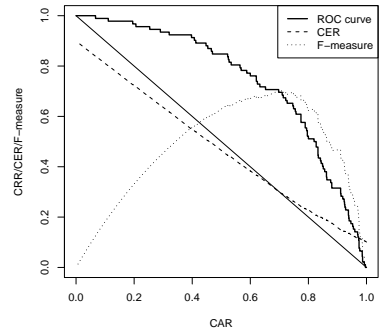}
  \caption{combination of the two best MI based CMs.}
  \label{ROCimicmi}
  \end{figure}

\vspace{-0.25cm}
We linearly combined the scores assigned to each word by different confidence measures to produce a new score. The weights are optimised by the perceptron algorithm on a small corpus (600 pairs of sentences), and tested on a different corpus (219 pairs of sentences). Figure \ref{ROCimicmi} shows the performances of the classifier resulting from the linear combination of the best previously presented intra-MI based classifier and the best inter-MI based one.

The combination of the two yields interesting improvement, especially in terms of error rate. The perceptron gave more weights to the inter-MI based scores, but that is because these scores are generally lower and does not mean that this measure is more significant. On the other hand, combining these two confidence measures with n-grams and linguistic features based ones did not bring any improvement over our test corpus.

\vspace{-0.4cm}
\section{\uppercase{Discussion and Conclusion}}
\vspace{-0.25cm}

In this article, we present confidence scores that showed interesting discriminating power. We summarised the results obtained by the best different estimators (in terms of F-measure) in Table \ref{tableBestCMs}. For comparison Blatz et al. obtain in \cite{blatz2003cem} a CER of 29.2\% by combining two different word posterior probability estimates (with and without alignment) and the translation probabilities from IBM-1 model.

\begin{table}[ht!]
\centering
\scriptsize
\begin{tabular}{|l|c|c|c|c|c|}
\hline
  & threshold & CER & CAR & CRR & F-measure \\ \hline
 intra-MI & $3.6\cdotp 10^{-5}$ & 0.383 & 0.600 & 0.760  & 0.700 \\ \hline
 inter-MI & 0.0008 & 0.368 & 0.620 & 0.724  & 0.668 \\ \hline
 4-gram model & 0.134 &  0.377 & 0.619 & 0.653 & 0.636 \\ \hline
 linguistic 6-grams & 0.188 & 0.422 & 0.578 & 0.574 & 0.576 \\ \hline
 combined MI & $9\cdotp 10^{-5}$ & 0.251 & 0.759 & 0.663 & 0.708 \\ \hline
\end{tabular}
\caption{Performances of the best classifiers.}
\label{tableBestCMs}
\end{table}

The settings used by the best intra-MI based confidence measures were the following: tool words were ignored, no normalisation was applied, and words positions were not taken into account. For the best inter-MI based CMs, tool words were not taken into account, no normalisation, and triggering was allowed within a centred window of width 9 (maximal ``distortion'' of 4). From these figures we can tell that the best MI-based confidence measures outperform significantly the other CMs presented here, especially when used in combination. Note however that the best classifiers in terms of F-measure are not necessarily the best ones with regard to other metrics, for example CER.

  \subsection{Application of Confidence Measures}
\vspace{-0.25cm}
  While they were not investigated in this article, we can imagine several applications to confidence estimation, beside manual correction of erroneous words: \textbf{pruning or reranking of the n-best list} according to the confidence score, \textbf{generation of new hypothesis} by recombining parts of different candidates having high scores, or \textbf{discriminating training} by tuning the parameters to optimise the separation between sentences (or words, or phrases) having a high confidence score (hopefully they are correct translations) and sentences having a low one.

  \subsection{Prospects}
\vspace{-0.25cm}
  We plan to go further in our investigation on confidence measures for SMT: first, while the confidence measures we used take into account word insertion and word substitution, they do not directly take into account word deletion nor word order, and neither do our reference corpus (in which words are labelled as correct or incorrect, but missing words are not indicated). This serious drawback has to be addressed. Assigning confidence scores to alignment might help to this end. Second, we believe that in a context of phrase-based translation, phrase-level confidence estimation would be more appropriate. Also many features used in speech recognition or automatic translation could be used in confidence estimation: distant models, word alignment, word spotting, etc... Another problem is the fusion of different classifiers. We use a very simple single layer perceptron, but many solutions have been proposed in literature to achieve more appropriate merging. Finally, progress could be made on classifiers' evaluation: because classifying a word as correct or incorrect is a very difficult task even for a human translator, and because the results of such a task may vary according to the translator or worse, vary along time for a given translator, we should combine different human-generated references.



\vspace{-0.4cm}
\renewcommand{\baselinestretch}{0.98}
\bibliographystyle{apalike}
{\small
\bibliography{confidence_measures-article}}
\renewcommand{\baselinestretch}{1}

\end {document}